\documentclass[runningheads]{llncs}
\usepackage[T1]{fontenc}
\usepackage{graphicx}
\usepackage[
backend=biber,
style=numeric,
sorting=none
]{biblatex}

\let\oldthebibliography\thebibliography
\renewcommand{\thebibliography}[1]{%
  \oldthebibliography{#1}%
  \fontsize{8}{9.5}\selectfont%
  \setlength{\itemsep}{0pt plus 0.5pt}%
  \setlength{\parsep}{0pt}%
  \setlength{\parskip}{0pt}%
}

\begin{document}
\vspace{-20pt}
\title{Multi-Agent Orchestration for High-Throughput Materials Screening on a Leadership-Class System}
\titlerunning{Multi-Agent Orchestration}
%
\author{Thang Duc Pham\inst{1}
\and
Harikrishna Tummalapalli\inst{2}
\and
Fakhrul Hasan Bhuiyan\inst{1}
\and
\'{A}lvaro V\'{a}zquez Mayagoitia\inst{1}
\and
Christine Simpson\inst{2}
\and
Riccardo Balin\inst{2}
\and
Venkatram Vishwanath\inst{2}
\and
Murat Keçeli\inst{1}}

%
\authorrunning{T. D. Pham et al.}
%
\institute{Computational Science (CPS) Division, Argonne National Laboratory, Lemont IL 60439, USA \and
Argonne Leadership Computing Facility (ALCF), Argonne National Laboratory, Lemont IL 60439, USA
}
%
\maketitle              
\vspace{-20pt}
\begin{abstract}
The integration of Artificial Intelligence (AI) with High-Performance Computing (HPC) is transforming scientific workflows from human-directed pipelines into adaptive systems capable of autonomous decision-making. 
Large language models (LLMs) play a critical role in autonomous workflows; however, deploying LLM-based agents at scale remains a significant challenge. 
Single-agent architectures and sequential tool calls often become serialization bottlenecks when executing large-scale simulation campaigns, failing to utilize the massive parallelism of exascale resources. 
To address this, we present a scalable, hierarchical multi-agent framework for orchestrating high-throughput screening campaigns. 
Our “planner–executor” architecture employs a central planning agent to dynamically partition workloads and assign subtasks to a swarm of parallel executor agents. 
All executor agents interface with a shared Model Context Protocol (MCP) server that orchestrates tasks via the Parsl workflow engine. 
To demonstrate this framework, we employed the open-weight gpt-oss-120b model to orchestrate a high-throughput screening of the Computation-Ready Experimental (CoRE) Metal-Organic Framework (MOF) database for atmospheric water harvesting.
The results demonstrate that the proposed agentic framework enables efficient and scalable execution on the Aurora supercomputer, with low orchestration overhead and high task completion rates.
This work establishes a flexible paradigm for LLM-driven scientific automation on HPC systems, with broad applicability to materials discovery and beyond.
\end{abstract}
\vspace{-20pt}
\section{Introduction}
Traditional High-Performance Computing (HPC) workflows excel at executing large, predefined ensembles of simulations at scale, but their largely static control flow makes it difficult to adapt dynamically to intermediate results and failures. 
Recent advances in large language models (LLMs) \cite{llmreview} offer a complementary capability: the ability to reason over complex scientific objectives and dynamically adapt execution strategies \cite{coscientist}. 
By coupling LLMs with external simulation tools, researchers can create autonomous, agentic workflows that translate high-level intent directly into executable computation.

However, scaling these agentic workflows to modern exascale systems presents a significant challenge. 
Many current agent-based systems rely on iterative reasoning loops in which tool calls are generated and executed sequentially.
In large-scale screening campaigns, this pattern can introduce substantial serialization overhead and prevent effective use of available HPC concurrency.
As a result, naive LLM-driven workflows may underperform relative to conventional workflows that are explicitly constructed for parallel execution.

To address this limitation, we introduce a scalable and hierarchical “planner–executor” multi-agent framework that decouples high-level reasoning from large-scale task execution. 
Implemented via LangGraph \cite{langgraph2024}, a central planner agent decomposes workloads while a dynamic swarm of executor agents processes tasks asynchronously. 
Tool execution is standardized through the Model Context Protocol (MCP) and integrated with the Parsl workflow engine \cite{parsl}. 
This separation of concerns allows the planner to focus on scientific intent and strategy, while Parsl manages resource allocation, fault tolerance, and large-scale concurrency. 
The resulting system enables LLM agents to orchestrate large simulation campaigns without direct exposure to low-level scheduling or execution details.

We demonstrate the capabilities of this approach using a high-throughput materials screening application focused on Metal-Organic Frameworks (MOFs) \cite{furukawa_chemistry_2013}.
MOFs have emerged as promising candidates for atmospheric water harvesting (AWH) \cite{yaghi2020}.
Identifying suitable MOFs for AWH requires evaluating adsorption behavior across specific humidity ranges, making it a computationally intensive screening problem over thousands of candidate materials \cite{moghadam2017}.
This makes AWH an ideal test case for evaluating intelligent, scalable workflows that can efficiently explore large materials spaces.

To this end, we employ the open-weight \texttt{gpt-oss-120b} model to orchestrate a high-throughput screening of the Computation-Ready Experimental (CoRE) MOF database \cite{coremofdb} on the Aurora supercomputer \cite{bertoni25aurora}. 
Our results show that the framework can coordinate large simulation campaigns with modest orchestration overhead while leveraging the scalable execution capabilities of Aurora through Parsl. 
More broadly, this study illustrates a practical route for combining LLM-based agents with workflow engines to support scientific automation on HPC systems.
\vspace{-7pt}
\section{Contributions}

This paper presents a scalable architecture for running LLM-driven scientific workflows on leadership-class HPC systems. 
Our key contributions are:

\textbf{Hierarchical multi-agent orchestration framework.}
We introduce a planner--executor architecture, implemented using LangGraph, that separates planning, execution, and analysis roles across multiple agents. This design enables high-level task decomposition, concurrent workflow generation, and improved modularity and traceability in complex scientific workflows. 
To support reproducibility and reuse, the framework is available as open-source software under a permissive license on GitHub.

\textbf{Asynchronous MCP–Parsl integration.}
We introduce a tool interface in which MCP tools generate Parsl applications rather than executing simulations directly. 
This enables asynchronous dispatch of large simulation ensembles while Parsl manages scheduling and fault tolerance.

\textbf{Demonstration on Aurora using an open-weight model.}
We demonstrate the framework in an end-to-end screening of the CoRE MOF database on the Aurora supercomputer using an open-weight model, showing that LLM-based orchestration can support large simulation campaigns with modest overhead relative to simulation time.

\vspace{-7pt}
\section{Related Work}
\vspace{-7pt}
\subsection{Agentic AI for Scientific Discovery}

LLMs are increasingly being used to build autonomous agents capable of interacting with external tools and executing multi-step scientific workflows. 
The ReAct framework \cite{react} introduced a paradigm in which models combine reasoning and tool execution, which has since been adopted by systems such as ChemCrow \cite{chemcrow}, ChemGraph \cite{chemgraph}, El Agente \cite{elagente}, and Cactus \cite{cactus}. 
These systems enable natural language interaction with simulation software, automating tasks such as input preparation, job execution, and analysis.

However, most existing implementations target interactive or small-scale workloads and do not explicitly address the concurrency requirements of large HPC campaigns. 
When thousands of simulations must be executed, sequential tool invocation can introduce significant serialization overhead.
Scaling agentic workflows to leadership-class HPC systems therefore remains an open challenge, particularly when thousands of independent simulations must be orchestrated efficiently.
Recent work has begun to explore distributed agent architectures in scientific settings \cite{academy}, highlighting the potential of multi-agent systems for coordinating more complex workloads.


\vspace{-7pt}
\subsection{Convergence of Agents and HPC}
Workflow management systems such as Parsl \cite{parsl}, FireWorks \cite{fireworks} and Balsam \cite{balsam} provide scalable infrastructure for executing large scientific workflows on HPC systems, providing abstractions for defining and executing distributed workflows while handling resource management, fault tolerance, and data dependencies. 
Parsl, in particular, enables users to express workflows as Python-based Directed Acyclic Graphs (DAGs) and has been shown to scale to exascale systems \cite{mprot_dpo}.

Despite their scalability, traditional WMS are largely static: workflow structure and execution logic are defined \textit{a priori} by human developers. 
They lack the semantic reasoning capabilities needed to adapt execution strategies dynamically based on intermediate results or evolving scientific objectives. 
Integrating LLM-based reasoning with robust workflow managers therefore represents a promising path toward dynamic, adaptive scientific workflows.

Recent efforts have attempted to bridge the gap between AI reasoning and HPC execution. 
The Colmena framework \cite{colmena} introduced the concept of ``Thinker'' and ``Doer'' agents, separating the inference tasks from the simulation tasks. 
This allows for interleaving AI inference with simulation ensembles, primarily for active learning applications.
Similarly, rapid prototyping of agents using LangChain has been explored in conjunction with Parsl to enable tool-use on supercomputers \cite{langchain_parsl}. 
Our work builds on this emerging intersection, but focuses specifically on hierarchical multi-agent orchestration and on translating agent-generated tool requests into asynchronously dispatched simulation ensembles through MCP and Parsl.

\vspace{-7pt}
\section{Methodology}



\vspace{-7pt}
\subsection{Model Context Protocol}\label{MCP}

We used the Model Context Protocol (MCP) \cite{mcp} to connect the LLM agents with external tools. 
Two MCP servers were implemented: (1) a Data Tool server for aggregating simulation outputs and ranking materials, and (2) a Chemistry server for launching GCMC simulations.

The Chemistry server exposes tools that generate Parsl applications corresponding to individual or ensemble simulation tasks. 
These tasks are submitted to a co-located Parsl manager, which schedules them across compute nodes. 
All tools are implemented asynchronously, enabling multiple executor agents to request concurrent simulations without blocking.

\vspace{-7pt}
\subsection{Grand Canonical Monte Carlo Simulations}\label{GCMC}
To demonstrate the workflow, we performed GCMC simulations using the gRASPA package \cite{zhaoli2024}. 
Each simulation executed on one tile of the Intel Data Center Max 1550 GPU. 
Framework atoms were modeled with the Universal Force Field (UFF) \cite{rappe1992}, while adsorbates were represented using standard models (TIP4P for H$_2$O \cite{jorgensen1983} and TraPPE for CO$_2$ and N$_2$ \cite{Siepmann2001-ff}). 
Framework charges were assigned using PACMOF2 \cite{pham2024}. 
Standard Lennard-Jones and Ewald summation methods were used to compute intermolecular interactions.

The MOF structures were obtained from the CoRE MOF 2025 database \cite{coremofdb}, consisting of 5,591 all-solvent-removed and computation-ready structures hosted by the Cambridge Crystallographic Data Centre \cite{ccdc_host}.
The simulation settings provide a realistic benchmark for evaluating the orchestration framework, rather than a definitive assessment of MOF performance for atmospheric water harvesting.

\vspace{-7pt}
\subsection{Agentic Workflow}
\begin{figure*}[t]
    \centering
    \includegraphics[width=0.9\textwidth]{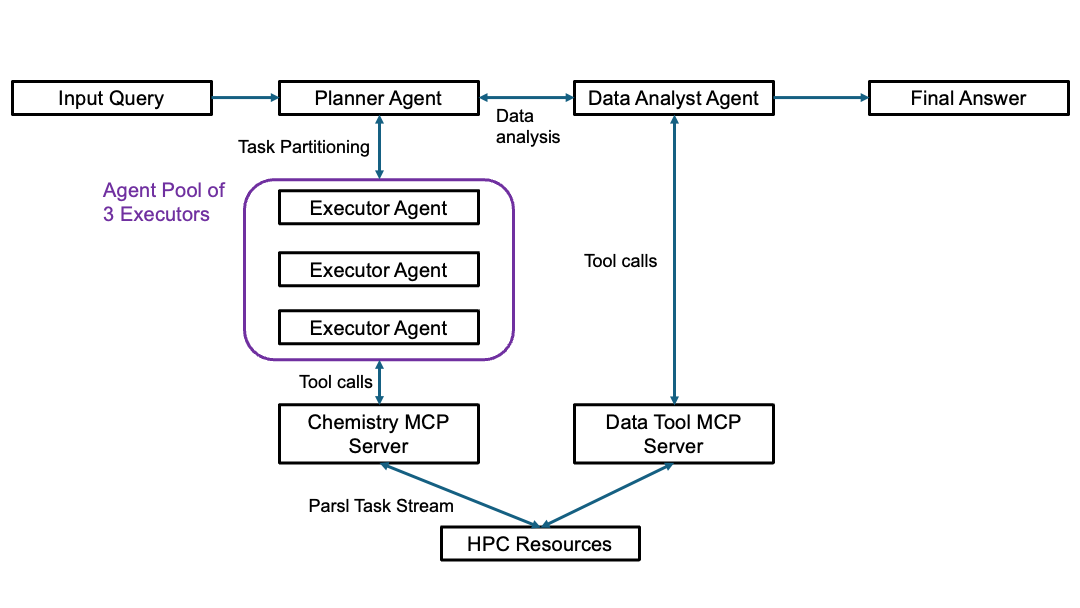}
    \vspace{-3mm}
    \caption{Schematic of the scalable multi-agent orchestration architecture, with a central planner agent, dynamically allocated number of executor agents, a data analyst agent and MCP servers.}
    \vspace{-6mm}
    \label{fig:workflow}
\end{figure*}

The architecture shown in Figure \ref{fig:workflow} extends our previously introduced \textit{ChemGraph} system \cite{chemgraph}, which enables LLM agents to orchestrate molecular simulation workflows through natural language interaction. 
While ChemGraph focuses on automating simulation setup, execution, and analysis, the present work extends the framework to support scalable orchestration of large simulation campaigns on HPC systems.

The system consists of a planner agent, a pool of executor agents, a data analyst agent, and two MCP servers.
Given a user query, the planner agent decomposes the scientific objective into independent simulation tasks and distributes them across the executor pool. 
Executor agents invoke simulation tools exposed by the Chemistry MCP server to launch atomistic simulations through Parsl. 
The number of executors scales dynamically based on the workload generated by the planner agent.

After the simulations are completed, the results are forwarded to the data analyst agent, which aggregates the outputs using tools provided by the Data Tool MCP server and produces the final response to the user query.
\subsection{Experimental Setup}

Experiments were conducted on the Aurora supercomputer at the Argonne Leadership Computing Facility (ALCF). 
Agent reasoning was performed using the gpt-oss-120b model deployed on the ALCF inference endpoints \cite{alcf_endpoint}.

The workflow was initiated using the structured prompt below, which defines the scientific objective and simulation parameters.
\vspace{-7pt}
\begin{quote}
    \textbf{Role:} You are coordinating a delegated workflow.
    
    \textbf{Data Source:} The CoRE MOF database is located at: 
    
    \texttt{/path/to/coremof-database}
    
    \textbf{Objective:} Identify the top 20\% best-performing MOFs for atmospheric water harvesting by calculating the working capacity between adsorption and desorption conditions.
    
    \textbf{Simulation Parameters:}
    \begin{itemize}
        \item Temperature: 298 K
        \item Water Saturation Pressure at 298K: 3200 Pa
        \item Adsorption condition: 60\% relative humidity 
        \item Desorption condition: 10\% relative humidity
        \item Duration: 2,000,000 (2 million) cycles per run
    \end{itemize}
\end{quote}

\vspace{-7pt}
\section{Results and Discussion}
\vspace{-7pt}
\subsection{Agentic Workflow Output Demonstration}

\begin{figure}
\centering
\includegraphics[width=0.9\textwidth]{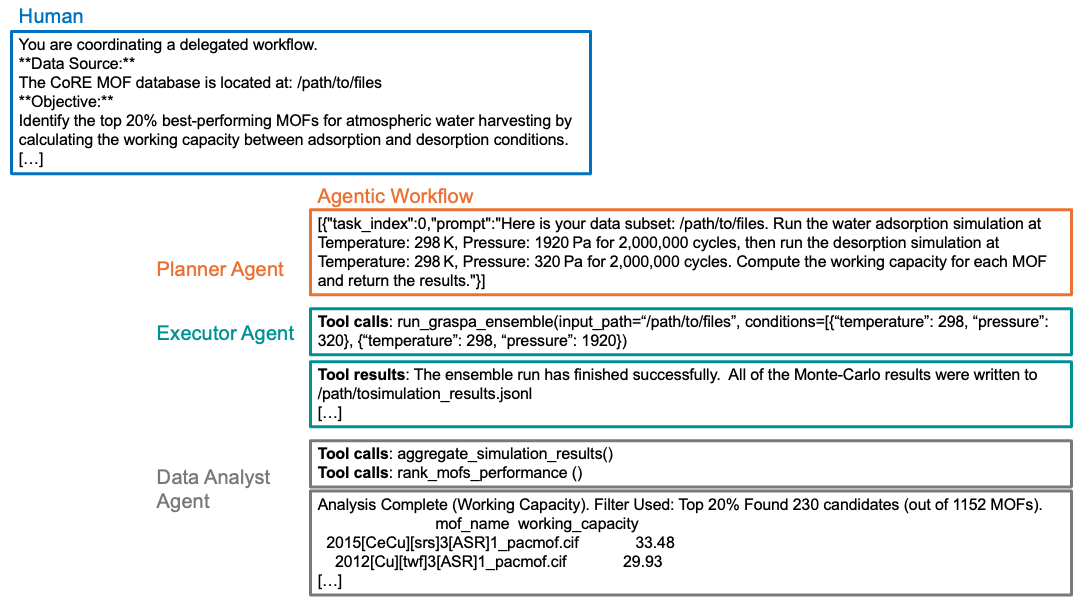}
    \vspace{-3mm}
\caption{Representative outputs from an agentic workflow used for high-throughput screening of 1,152 MOFs for atmospheric water harvesting. The figure illustrates the end-to-end interaction, including the human query, the planner agent’s task decomposition, the executor agent’s tool invocation and corresponding outputs, and the data analyst agent’s post-processing steps and final response. For clarity, both the user query and agent outputs are abbreviated using [...].}
    \vspace{-6mm}
    
\label{fig:example_1_exec}
\end{figure}

Figure \ref{fig:example_1_exec} demonstrates the representative input and output of our agentic workflow. Starting from a human natural language query, the planner agent interprets the scientific objective and decomposes it into structured, executable tasks. The task is then dispatched to the executor agent, which invokes simulation tools and records both the tool calls and their returned outputs. The resulting simulation data (saved as a JSONL file) are then processed by a data analyst agent, which performs post-processing, aggregation, and ranking to generate a final, user-facing response. 

This example highlights the transparent step-by-step reasoning process enabled by the agentic design, where intermediate decisions and tool interactions are explicitly exposed. This traceability is critical for debugging, validation, and scaling high-throughput scientific workflows on HPC systems, particularly when screening large design spaces.

\vspace{-7pt}
\subsection{High-throughput Screening Results}
\begin{figure}
\centering
\includegraphics[width=0.8\textwidth]{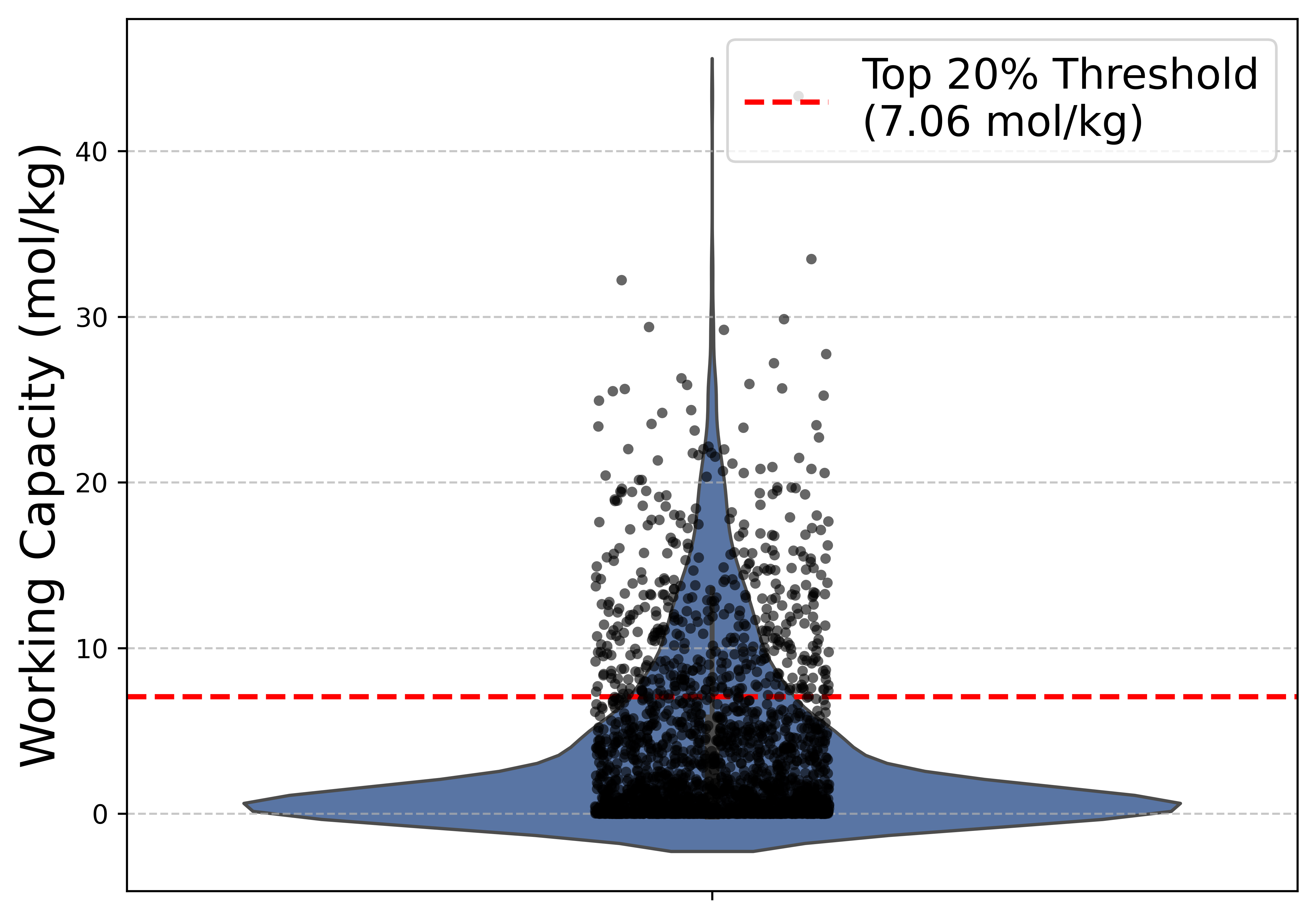}
    \vspace{-3mm}
    \caption{Distribution of working capacities of water for the screened 2,304 Metal-Organic Frameworks (MOFs), calculated between 1920 Pa (adsorption) and 320 Pa (desorption) at 298 K, from the 256 nodes weak-scaling run. The violin plot illustrates the probability density of the dataset, while the overlaid strip plot represents individual MOF candidates. The red-dashed line marks the 80th percentile (top 20\% cutoff).}
        \vspace{-6mm}
\label{fig:htsresult}
\end{figure}

Building on the demonstration illustrated in Figure \ref{fig:example_1_exec}, we plotted the water working capacities of the 2,304 MOFs, defined as the difference in the amount of water adsorbed at the adsorption condition (298 K, 1920 Pa) and desorption condition (298 K, 320 Pa). Figure \ref{fig:htsresult} shows the distribution of working capacities computed by the agentic workflow, verifying the agent's ability to manage thousands of concurrent GCMC simulations and aggregate the resulting data without human intervention.

The screening analysis of 2,304 MOFs reveals a highly skewed distribution, with the majority of candidates exhibiting working capacities below 1.0 mol/kg, making them unsuitable for practical atmospheric water harvesting. In contrast, the top 20\% of candidates demonstrated high performance, with working capacities achieving 7.06 mol/kg. 
While the identification of a definitive optimal material requires further consideration of hydrothermal stability and specific cycling conditions, the physical trends captured here provide a useful screening-level assessment. 
More importantly, these results demonstrate that agentic workflows can streamline high-throughput research by allowing researchers to execute large-scale simulation campaigns through a simple natural language query.

\vspace{-7pt}
\subsection{Multi-Objective Workflow}

Scientific screening campaigns often require evaluating candidate materials against multiple objectives simultaneously. 
To test this capability, we issued a natural language query instructing the agentic system to perform a multi-objective screening.
The request asked the system to evaluate three adsorption scenarios: water adsorption at 60\% RH and 298 K, CO$_2$ adsorption at 0.15 bar and 298 K, and N$_2$ adsorption at 1 bar and 77 K, and then identify the top 20\% performing MOFs for each application.
\vspace{-7pt}
\begin{quote}
    \textbf{Role:} You are coordinating a delegated workflow.
    
    \textbf{Data Source:} The CoRE MOF database is located at: \texttt{/path/to/coremof-database}
    
    \textbf{Objective:} I want to run the following simulations:
    \begin{itemize}
    \item Water adsorption at 60\% RH, 298K
    \item CO$_2$ adsorption at 0.15 bar, 298K
    \item N$_2$ adsorption at 1 bar, 77K
    \item Identify the top 20\% best-performing MOFs for each application.
    \end{itemize}

    \textbf{Simulation Parameters:}
    \begin{itemize}
        \item Water Saturation Pressure at 298K: 3200 Pa
        \item Duration: 50,000 cycles per simulation
    \end{itemize}
\end{quote}

The planner agent interpreted this high-level query and decomposed it into three independent simulation tasks corresponding to the specified adsorption conditions. 
These tasks were distributed across three executor agents, which launched the simulations in parallel using the Chemistry MCP tools. 
After completion, the data analyst agent aggregated the results and ranked the materials to identify the top 20\% performing MOFs for each objective.

This experiment highlights a key advantage of the architecture: multi-objective workflows can be generated dynamically from natural language instructions without modifying the underlying codebase. 
Because the planner agent translates user intent into executable tasks at runtime, new screening campaigns with different properties, conditions, or objectives can be expressed directly through the interface while reusing the same execution framework.

\vspace{-7pt}
\subsection{Scaling}

\vspace{-7pt}
\subsubsection{Weak Scaling}

To test the throughput stability, we performed weak scaling experiments in which the number of MOF structures increases proportionally with the compute resources. We used a fixed workload of 9 MOFs per node (18 simulations per node).

Although the GCMC simulations used a fixed number of cycles, individual simulation times fluctuated significantly depending on the size of the simulation box (MOF crystal structure) and the number of water molecules adsorbed. Observed simulation times ranged from 1,600 seconds to 4,400 seconds. To investigate the impact of structural complexity on scaling, we employed two sampling strategies: (1) Random Sampling: MOFs were selected randomly for each run, and
(2) Nested Sampling: A cumulative strategy where the set of MOFs used in smaller scale runs is retained and expanded upon for larger scale runs.

\begin{figure*}[t]
\centering
\includegraphics[width=\textwidth]{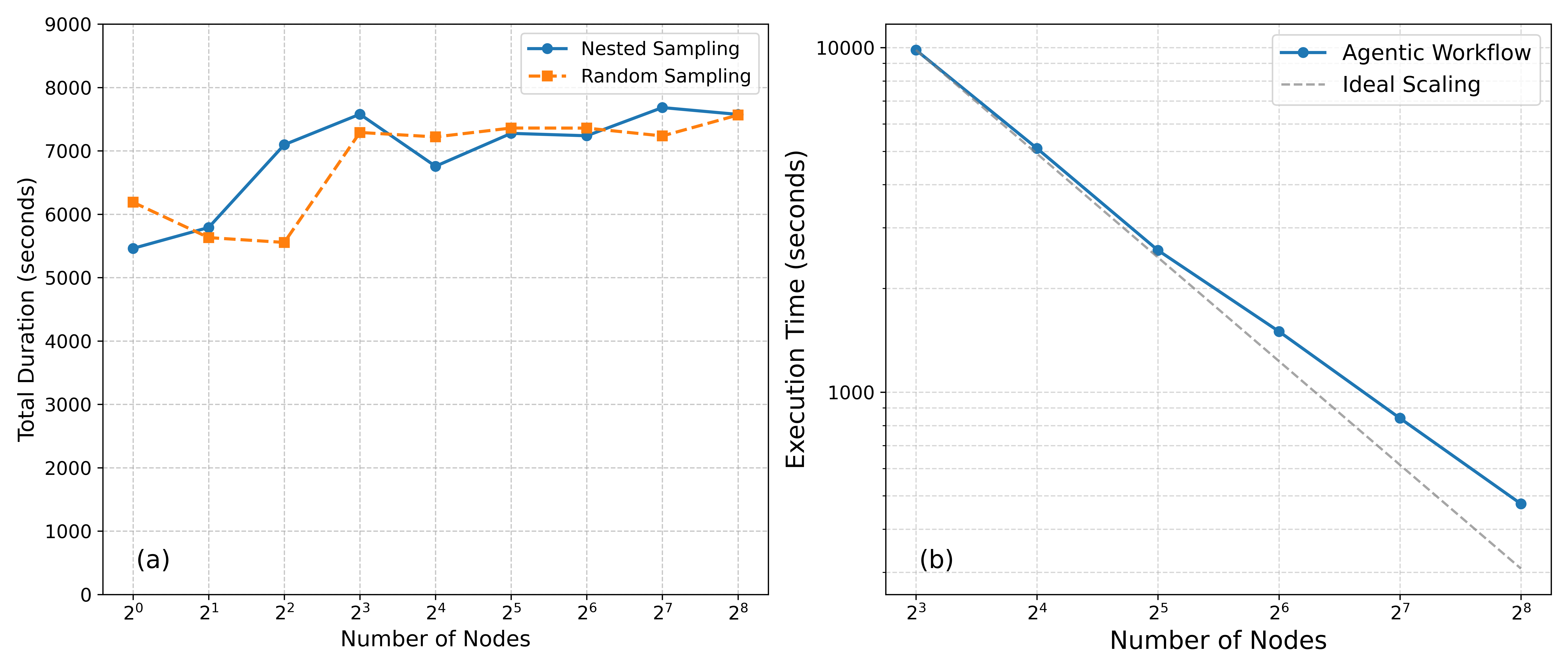}
\caption{Scaling performance of the multi-agent orchestration workflow. (a) Weak scaling with a constant workload of 9 MOFs per node across 1 to 256 nodes. (b) Strong scaling with a fixed workload of 5,591 MOFs (11,182 simulations) across 8 to 256 compute nodes.}
\label{fig:scaling}
\end{figure*}

Figure \ref{fig:scaling}a compares the weak scaling performance of these two strategies. As the system scales from 1 to 256 nodes, the total workflow durations for both strategies remain relatively stable between 5,500 and 8,000 seconds and indicate effective weak scaling.

\subsubsection{Strong scaling}

To evaluate the strong scaling performance, we ran the agentic workflow with the complete dataset of 5,591 MOFs while increasing compute resources from 8 to 256 nodes. For these experiments, the GCMC simulation was adjusted to 50,000 cycles to facilitate rapid throughput. 
This reduced number of cycles was used only for the scaling benchmark, whereas the 2,000,000-cycle setting for the weak-scaling runs was used for production calculations. 
The choice of 50,000 cycles was made so that the full 5,591-MOF workload could be completed on 8 Aurora nodes within the wall-time limit.
Figure \ref{fig:scaling}b illustrates the execution time as a function of node count. The workflow demonstrates near-linear speedup, tracking the ideal strong scaling curve from 8 to 32 nodes. At higher node counts (128 and 256), the system maintains a significant reduction in time-to-solution, though strong scaling efficiency drops down to 64.9\%.


\vspace{-7pt}
\subsection{Agentic Framework Evaluation}

In this work, we employed gpt-oss-120b, an open-weight model, to drive the agentic workflow. 
While proprietary models often perform best in agentic benchmarks \cite{berkerley_toolcall, chemgraph}, utilizing gpt-oss-120b allows users to self-host both the model and workflow logic. 
This is crucial for scalability, because as workflows grow in complexity, the token costs associated with proprietary APIs can become a significant financial burden for researchers.

In terms of performance, we observed a total agentic overhead (comprising API calls and LangGraph orchestration, excluding simulation timing) of approximately 60 to 90 seconds across independent runs. While this provides a baseline for the orchestration cost, it is important to note that this overhead is variable and highly dependent on model inference throughput and API latency.

In terms of execution reliability, we occasionally observed failures in gpt-oss-120b's tool-calling capabilities, where invalid generated arguments led to workflow termination. To mitigate this, failed runs were restarted from the initial query. From the weak and strong scaling runs, we performed a total of 25 experiments. We observed 4 failures, resulting in an overall success rate of 84\%.

\vspace{-7pt}
\section{Conclusions}

A major barrier to utilizing leadership-class HPC systems is the steep learning curve associated with environment management, task orchestration, and job scheduling. 
Our results demonstrate that an agentic workflow can bridge the gap between natural language requests and HPC execution. 
By successfully interpreting high-level scientific queries, ranging from simple single-task requests to complex multi-objective screenings, the agentic framework presented here allows researchers to perform a large-scale simulation campaign using a simple prompt.

Single- and multi-objective tests demonstrate the flexibility of the framework, while the weak and strong scaling experiments highlight the scalability of our approach across diverse workflow settings. 
Even with the significant variance in simulation times caused by different MOFs, the workflow maintained stable throughput. 
This suggests that AI agents, when paired with a workflow manager like Parsl, are well-suited for high-throughput screening campaigns where the computational cost per candidate is non-uniform, a common scenario in materials discovery.

An important finding of this work is that gpt-oss-120b, an open-weight model, can support complex scientific workflows. 
The observed success rate of 84\% indicates a remaining reliability gap, but this gap is likely to narrow as open-weight models continue to improve. 
In addition to cost advantages, open-weight models enable self-hosting and provide greater control over privacy, deployment, and reproducibility.
It should also be noted that the framework itself is model-agnostic and can be used with either open-weight or frontier models.

\vspace{-7pt}
\section{Acknowledgment}
This work was supported by the Office of Science, U.S. Department of Energy, under contract DE-AC02-06CH11357. 
MK is supported by the U.S. Department of Energy (DOE), Office of Science, Office of Basic Energy Science, Division of Chemical Sciences, Geosciences, and Biosciences through Argonne National Laboratory under Contract No. DE-AC02-06CH11357.
A.V.M. was supported by the Office of Science, U.S. Department of Energy, under contract DE-AC02-06CH11357.
This research used resources of the Argonne Leadership Computing Facility, a U.S.
Department of Energy (DOE) Office of Science user facility at Argonne National Laboratory. 
This work leverages ALCF Inference Endpoints, which provide a robust API for LLM inference on ALCF HPC clusters via Globus Compute.
The authors thank Dr. Abhishek Bagusetty for assistance with gRASPA software.

\vspace{-7pt}
\section{Code Availability}
\vspace{-4pt}
ChemGraph is open-source and available on GitHub at 

\url{https://github.com/argonne-lcf/ChemGraph}.

\vspace{-7pt}
\renewcommand*{\bibfont}{\footnotesize}
\printbibliography

@article{yaghi2020,
	title = {Metal–Organic Frameworks for Water Harvesting from Air, Anywhere, Anytime},
	volume = {6},
	issn = {2374-7943},
	doi = {10.1021/acscentsci.0c00678},
	pages = {1348--1354},
	number = {8},
	journaltitle = {{ACS} Central Science},
	shortjournal = {{ACS} Cent. Sci.},
	author = {Xu, Wentao and Yaghi, Omar M.},
	date = {2020-08-26},
}

@article{moghadam2017,
	title = {Development of a Cambridge Structural Database Subset: A Collection of Metal–Organic Frameworks for Past, Present, and Future},
	volume = {29},
	issn = {0897-4756},
	doi = {10.1021/acs.chemmater.7b00441},
	pages = {2618--2625},
	number = {7},
	journaltitle = {Chemistry of Materials},
	shortjournal = {Chem. Mater.},
	author = {Moghadam, Peyman Z. and Li, Aurelia and Wiggin, Seth B. and Tao, Andi and Maloney, Andrew G. P. and Wood, Peter A. and Ward, Suzanna C. and Fairen-Jimenez, David},
	date = {2017-04-11},
}

@article{zhaoli2024,
	title = {Efficient Implementation of Monte Carlo Algorithms on Graphical Processing Units for Simulation of Adsorption in Porous Materials},
	volume = {20},
	issn = {1549-9618},
	doi = {10.1021/acs.jctc.4c01058},
	pages = {10649--10666},
	number = {23},
	journaltitle = {Journal of Chemical Theory and Computation},
	shortjournal = {J. Chem. Theory Comput.},
	author = {Li, Zhao and Shi, Kaihang and Dubbeldam, David and Dewing, Mark and Knight, Christopher and Vázquez-Mayagoitia, Álvaro and Snurr, Randall Q.},
	date = {2024-12-10},
}

@article{rappe1992,
author = {Rappe, A. K. and Casewit, C. J. and Colwell, K. S. and Goddard, W. A. III and Skiff, W. M.},
title = {UFF, a full periodic table force field for molecular mechanics and molecular dynamics simulations},
journal = {Journal of the American Chemical Society},
volume = {114},
number = {25},
pages = {10024-10035},
year = {1992},
doi = {10.1021/ja00051a040},
}

@article{jorgensen1983,
	title = {Comparison of simple potential functions for simulating liquid water},
	volume = {79},
	issn = {0021-9606},
	doi = {10.1063/1.445869},
	pages = {926--935},
	number = {2},
	journaltitle = {The Journal of Chemical Physics},
	author = {Jorgensen, William L. and Chandrasekhar, Jayaraman and Madura, Jeffry D. and Impey, Roger W. and Klein, Michael L.},
	date = {1983-07},
}

@article{pham2024,
	title = {Predicting Partial Atomic Charges in Metal–Organic Frameworks: An Extension to Ionic {MOFs}},
	volume = {128},
	doi = {10.1021/acs.jpcc.4c04879},
	pages = {17165--17174},
	number = {40},
	journaltitle = {The Journal of Physical Chemistry C},
	shortjournal = {J. Phys. Chem. C},
	author = {Pham, Thang D. and Joodaki, Faramarz and Formalik, Filip and Snurr, Randall Q.},
	date = {2024-10-10},
}

@article{Siepmann2001-ff,
author = {Potoff, Jeffrey J. and Siepmann, J. Ilja},
title = {Vapor–liquid equilibria of mixtures containing alkanes, carbon dioxide, and nitrogen},
journal = {AIChE Journal},
volume = {47},
number = {7},
pages = {1676-1682},
doi = {https://doi.org/10.1002/aic.690470719},
year = {2001}
}

@misc{react,
      title={ReAct: Synergizing Reasoning and Acting in Language Models}, 
      author={Shunyu Yao and Jeffrey Zhao and Dian Yu and Nan Du and Izhak Shafran and Karthik Narasimhan and Yuan Cao},
      year={2023},
      eprint={2210.03629},
      archivePrefix={arXiv},
      primaryClass={cs.CL},
      url={https://arxiv.org/abs/2210.03629}, 
}

@article{elagente,
	title = {El Agente: An autonomous agent for quantum chemistry},
	volume = {8},
	issn = {2590-2393},
	doi = {10.1016/j.matt.2025.102263},
	number = {7},
	journaltitle = {Matter},
	author = {Zou, Yunheng and Cheng, Austin H. and Aldossary, Abdulrahman and Bai, Jiaru and Leong, Shi Xuan and Campos-Gonzalez-Angulo, Jorge Arturo and Choi, Changhyeok and Ser, Cher Tian and Tom, Gary and Wang, Andrew and Zhang, Zijian and Yakavets, Ilya and Hao, Han and Crebolder, Chris and Bernales, Varinia and Aspuru-Guzik, Alán},
	date = {2025-07-02},
}

@article{cactus,
	title = {{CACTUS}: Chemistry Agent Connecting Tool Usage to Science},
	volume = {9},
	doi = {10.1021/acsomega.4c08408},
	pages = {46563--46573},
	number = {46},
	journaltitle = {{ACS} Omega},
	shortjournal = {{ACS} Omega},
	author = {{McNaughton}, Andrew D. and Sankar Ramalaxmi, Gautham Krishna and Kruel, Agustin and Knutson, Carter R. and Varikoti, Rohith A. and Kumar, Neeraj},
	date = {2024-11-19},
}

@misc{mcp,
      title={Model Context Protocol (MCP): Landscape, Security Threats, and Future Research Directions}, 
      author={Xinyi Hou and Yanjie Zhao and Shenao Wang and Haoyu Wang},
      year={2025},
      eprint={2503.23278},
      archivePrefix={arXiv},
      primaryClass={cs.CR},
      url={https://arxiv.org/abs/2503.23278}, 
}

@article{coscientist,
  title={Autonomous chemical research with large language models},
  author={Boiko, Daniil A and MacKnight, Robert and Kline, Ben and Gomes, Gabe},
  journal={Nature},
  volume={624},
  number={7992},
  pages={570--578},
  year={2023},
  publisher={Nature Publishing Group},
  doi={10.1038/s41586-023-06792-0}
}

@article{chemcrow,
  title={Augmenting large language models with chemistry tools},
  author={Bran, Andres M and Cox, Sam and White, Andrew D and Schwaller, Philippe},
  journal={Nature Machine Intelligence},
  volume={6},
  pages={525--535},
  year={2024},
  %publisher={Nature Publishing Group}
}

@article{chemgraph,
	title = {{ChemGraph} as an agentic framework for computational chemistry workflows},
	volume = {9},
	issn = {2399-3669},
	doi = {10.1038/s42004-025-01776-9},
	pages = {33},
	number = {1},
	journaltitle = {Communications Chemistry},
	shortjournal = {Communications Chemistry},
	author = {Pham, Thang D. and Tanikanti, Aditya and Keçeli, Murat},
	date = {2026-01-08},
}

@inproceedings{parsl,
  title={Parsl: Pervasive parallel programming in python},
  author={Babuji, Yadu and Woodard, Anna and Li, Zhuozhao and Katz, Daniel S and Clifford, Ben and Kumar, Rohan and Lacinski, Lukasz and Chard, Ryan and Wozniak, Justin M and Foster, Ian and others},
  booktitle={28th International Symposium on High-Performance Parallel and Distributed Computing (HPDC)},
  pages={25--36},
  year={2019}
}

@inproceedings{colmena,
  title={Colmena: Scalable steering of ensemble simulations with artificial intelligence},
  author={Ward, Logan and Blaiszik, Ben and Foster, Ian and Assary, Rajeev and Narayanan, Badri and Babuji, Yadu},
  booktitle={Proceedings of the International Conference for High Performance Computing, Networking, Storage and Analysis (SC21)},
  pages={1--12},
  year={2021}
}

@article{fireworks,
  title={FireWorks: a dynamic workflow system designed for high-throughput applications},
  author={Jain, Anubhav and Ong, Shyue Ping and Chen, Wei and Medasani, Bharat and Qu, Xiaohui and Kocher, Michael and Brafman, Marsha and Petasch, Guido and Gunter, Dan and Hautier, Geoffroy and others},
  journal={Concurrency and Computation: Practice and Experience},
  volume={27},
  number={17},
  pages={5037--5059},
  year={2015},
}

@inproceedings{balsam,
author={Salim, Michael and Uram, Thomas and Childers, J. Taylor and Vishwanath, Venkatram and Papka, Michael},
  booktitle={2019 IEEE/ACM 1st Annual Workshop on Large-scale Experiment-in-the-Loop Computing (XLOOP)}, 
  title={Balsam: Near Real-Time Experimental Data Analysis on Supercomputers}, 
  year={2019},
  volume={},
  number={},
  pages={26-31},
  doi={10.1109/XLOOP49562.2019.00010}}

@inproceedings{alcf_endpoint,
author = {Tanikanti, Aditya and C\^{o}t\'{e}, Benoit and Guo, Yanfei and Chen, Le and Saint, Nickolaus and Chard, Ryan and Raffenetti, Ken and Thakur, Rajeev and Uram, Thomas and Foster, Ian and Papka, Michael E. and Vishwanath, Venkatram},
title = {FIRST: Federated Inference Resource Scheduling Toolkit for Scientific AI Model Access},
year = {2025},
isbn = {9798400718717},
%publisher = {Association for Computing Machinery},
address = {New York, NY, USA},
doi = {10.1145/3731599.3767346},
abstract = {We present the Federated Inference Resource Scheduling Toolkit (FIRST), a framework enabling Inference-as-a-Service across distributed High-Performance Computing (HPC) clusters. FIRST provides cloud-like access to diverse AI models, like Large Language Models (LLMs), on existing HPC infrastructure. Leveraging Globus Auth and Globus Compute, the system allows researchers to run parallel inference workloads via an OpenAI-compliant API on private, secure environments. This cluster-agnostic API allows requests to be distributed across federated clusters, targeting numerous hosted models. FIRST supports multiple inference backends (e.g., vLLM), auto-scales resources, maintains "hot" nodes for low-latency execution, and offers both high-throughput batch and interactive modes. The framework addresses the growing demand for private, secure, and scalable AI inference in scientific workflows, allowing researchers to generate billions of tokens daily on-premises without relying on commercial cloud infrastructure.},
booktitle = {Proceedings of the SC '25 Workshops of the International Conference for High Performance Computing, Networking, Storage and Analysis},
pages = {52–60},
numpages = {9},
keywords = {Inference as a Service, High Performance Computing, Job Schedulers, Large Language Models, Globus, Scientific Computing},
location = {
},
series = {SC Workshops '25}
}

@article{coremofdb,
	title = {{CoRE} {MOF} {DB}: A curated experimental metal-organic framework database with machine-learned properties for integrated material-process screening},
	volume = {8},
	issn = {2590-2385},
	doi = {https://doi.org/10.1016/j.matt.2025.102140},
	pages = {102140},
	number = {6},
	journaltitle = {Matter},
	author = {Zhao, Guobin and Brabson, Logan M. and Chheda, Saumil and Huang, Ju and Kim, Haewon and Liu, Kunhuan and Mochida, Kenji and Pham, Thang D. and {Prerna} and Terrones, Gianmarco G. and Yoon, Sunghyun and Zoubritzky, Lionel and Coudert, François-Xavier and Haranczyk, Maciej and Kulik, Heather J. and Moosavi, Seyed Mohamad and Sholl, David S. and Siepmann, J. Ilja and Snurr, Randall Q. and Chung, Yongchul G.},
	date = {2025},
}

@misc{ccdc_host,
  title        = {{Computation Ready Metal--Organic Frameworks (CoRE MOF) Database}},
  author       = {{Cambridge Crystallographic Data Centre (CCDC)}},
  year         = {2025},
  note         = {Accessed: 2025-11-01}
}

@inproceedings{
berkerley_toolcall,
title={The Berkeley Function Calling Leaderboard ({BFCL}): From Tool Use to Agentic Evaluation of Large Language Models},
author={Shishir G Patil and Huanzhi Mao and Fanjia Yan and Charlie Cheng-Jie Ji and Vishnu Suresh and Ion Stoica and Joseph E. Gonzalez},
booktitle={Forty-second International Conference on Machine Learning},
year={2025},
url={https://openreview.net/forum?id=2GmDdhBdDk}
}

@misc{langgraph2024,
  title        = {LangGraph: A Framework for Building Stateful, Multi-Actor Applications with LLMs},
  author       = {{LangChain, Inc.}},
  year         = {2025},
  howpublished = {\url{https://github.com/langchain-ai/langgraph}},
}

@article{llmreview,
author = {Naveed, Humza and Khan, Asad Ullah and Qiu, Shi and Saqib, Muhammad and Anwar, Saeed and Usman, Muhammad and Akhtar, Naveed and Barnes, Nick and Mian, Ajmal},
title = {A Comprehensive Overview of Large Language Models},
year = {2025},
issue_date = {October 2025},
%publisher = {Association for Computing Machinery},
address = {New York, NY, USA},
volume = {16},
number = {5},
issn = {2157-6904},
doi = {10.1145/3744746},
abstract = {Large Language Models (LLMs) have recently demonstrated remarkable capabilities in natural language processing tasks and beyond. This success of LLMs has led to a large influx of research contributions in this direction. These works encompass diverse topics such as architectural innovations, better training strategies, context length improvements, fine-tuning, multimodal LLMs, robotics, datasets, benchmarking, efficiency, and more. With the rapid development of techniques and regular breakthroughs in LLM research, it has become considerably challenging to perceive the bigger picture of the advances in this direction. Considering the rapidly emerging plethora of literature on LLMs, it is imperative that the research community is able to benefit from a concise yet comprehensive overview of the recent developments in this field. This article provides an overview of the literature on a broad range of LLM-related concepts. Our self-contained comprehensive overview of LLMs discusses relevant background concepts along with covering the advanced topics at the frontier of research in LLMs. This review article is intended to provide not only a systematic survey but also a quick, comprehensive reference for the researchers and practitioners to draw insights from extensive, informative summaries of the existing works to advance the LLM research.},
journal = {ACM Trans. Intell. Syst. Technol.},
month = aug,
articleno = {106},
numpages = {72},
keywords = {Large Language Models, LLMs, chatGPT, Augmented LLMs, Multimodal LLMs, LLM training, LLM Benchmarking}
}

@misc{academy,
      title={Empowering Scientific Workflows with Federated Agents}, 
      author={Alok Kamatar and J. Gregory Pauloski and Yadu Babuji and Ryan Chard and Mansi Sakarvadia and Daniel Babnigg and Kyle Chard and Ian Foster},
      year={2026},
      eprint={2505.05428},
      archivePrefix={arXiv},
      primaryClass={cs.MA},
      url={https://arxiv.org/abs/2505.05428}, 
}

@inproceedings{langchain_parsl,
author = {Ma, Heng and Brace, Alexander and Siebenschuh, Carlo and Foster, Ian and Ramanathan, Arvind},
title = {LangChain-Parsl: Connect Large Language Model Agents to High Performance Computing Resource},
year = {2025},
isbn = {9798400718717},
%ublisher = {Association for Computing Machinery},
address = {New York, NY, USA},
doi = {10.1145/3731599.3767349},
abstract = {Large Language Models (LLMs) can improve performance in answering questions beyond their contextual understanding by running external tools, such as a calculator for arithmetics, an online query for real-time weather, et al. For scientific applications, this enables the LLM to perform and analyze simulation runs for more accurate answers. However, the increasing scale of scientific computing requires high-performance computers (HPCs), which are managed by job schedulers. In this work, we implemented Parsl to the LangChain tool calling to bridge the gap between the LLM agent and the HPC resource. Two implementations were set up and tested on a local Nvidia GPU workstation and the Polaris/ALCF HPC system. The first setup was implemented by modifying the LangChain tool calling, which converts the LangChain tool calls to Parsl functions and queues them to the Parsl workers for parallel execution. The second approach was achieved by designing a Parsl ensemble function as an LLM tool, which performed parallel tasks. With these implementations, the LLM agent workflow was prompted to run molecular dynamics simulations, with different protein structures and simulation conditions. The results show that our Parsl implementations enable parallel execution of scientific tools that invoked by LLM agents on both local GPU workstations and HPC platforms.},
booktitle = {Proceedings of the SC '25 Workshops of the International Conference for High Performance Computing, Networking, Storage and Analysis},
pages = {78–85},
numpages = {8},
keywords = {Large language model (LLM), agentic workflow, high performance computer, molecular dynamics simulation, LangChain/LangGraph, Parsl},
location = {
},
series = {SC Workshops '25}
}

@inproceedings{mprot_dpo,
author = {Dharuman, Gautham and Hippe, Kyle and Brace, Alexander and Foreman, Sam and Hatanp\"{a}\"{a}, V\"{a}in\"{o} and Sastry, Varuni K. and Zheng, Huihuo and Ward, Logan and Muralidharan, Servesh and Vasan, Archit and Kale, Bharat and Mann, Carla M. and Ma, Heng and Cheng, Yun-Hsuan and Zamora, Yuliana and Liu, Shengchao and Xiao, Chaowei and Emani, Murali and Gibbs, Tom and Tatineni, Mahidhar and Canchi, Deepak and Mitchell, Jerome and Yamada, Koichi and Garzaran, Maria and Papka, Michael E. and Foster, Ian and Stevens, Rick and Anandkumar, Anima and Vishwanath, Venkatram and Ramanathan, Arvind},
title = {MProt-DPO: Breaking the ExaFLOPS Barrier for Multimodal Protein Design Workflows with Direct Preference Optimization},
year = {2024},
isbn = {9798350352917},
%publisher = {IEEE Press},
doi = {10.1109/SC41406.2024.00013},
abstract = {We present a scalable, end-to-end workflow for protein design. By augmenting protein sequences with natural language descriptions of their biochemical properties, we train generative models that can be preferentially aligned with protein fitness landscapes. Through complex experimental- and simulation-based observations, we integrate these measures as preferred parameters for generating new protein variants and demonstrate our workflow on five diverse supercomputers. We achieve >1 ExaFLOPS sustained performance in mixed precision on each supercomputer and a maximum sustained performance of 4.11 ExaFLOPS and peak performance of 5.57 ExaFLOPS. We establish the scientific performance of our model on two tasks: (1) across a predetermined benchmark dataset of deep mutational scanning experiments to optimize the fitness-determining mutations in the yeast protein HIS7, and (2) in optimizing the design of the enzyme malate dehydrogenase to achieve lower activation barriers (and therefore increased catalytic rates) using simulation data. Our implementation thus sets high watermarks for multimodal protein design workflows.},
booktitle = {Proceedings of the International Conference for High Performance Computing, Networking, Storage, and Analysis},
articleno = {7},
numpages = {13},
keywords = {AI, HPC, Large language models, protein design},
location = {Atlanta, GA, USA},
series = {SC '24}
}

@inproceedings{bertoni25aurora,
author = {Bertoni, Colleen and Kwack, JaeHyuk and Applencourt, Thomas and Bagusetty, Abhishek and Ghadar, Yasaman and Homerding, Brian and Knight, Christopher and Luo, Ye and Thavappiragasam, Mathialakan and Tramm, John and Rangel, Esteban and Unnikrishnan, Umesh and Williams, Timothy J. and Parker, Scott},
title = {Early Application Experiences on Aurora at ALCF: Moving From Petascale to Exascale Systems},
year = {2025},
isbn = {9798400713286},
publisher = {Association for Computing Machinery},
address = {New York, NY, USA},
doi = {10.1145/3725789.3725791},
booktitle = {Proceedings of the Cray User Group},
pages = {12–23},
numpages = {12},
location = {
},
series = {CUG '24}
}

@article{furukawa_chemistry_2013,
	title = {The Chemistry and Applications of Metal-Organic Frameworks},
	volume = {341},
	doi = {10.1126/science.1230444},
	pages = {1230444},
	number = {6149},
	journaltitle = {Science},
	shortjournal = {Science},
	publisher = {American Association for the Advancement of Science},
	author = {Furukawa, Hiroyasu and Cordova, Kyle E. and O’Keeffe, Michael and Yaghi, Omar M.},
	date = {2013-08-30},
}
\vspace{-7pt}

\end{document}